\documentclass[11pt]{article}

\usepackage{dialogue}
\usepackage{booktabs}
\usepackage{ifxetex}
\ifxetex
    \usepackage{fontspec}
    \setromanfont{Times New Roman}
\else
  \usepackage{cmap}
  \usepackage[T2A,T1]{fontenc}
  \usepackage[utf8]{inputenc}
  \usepackage{times}
  \usepackage{latexsym}
  \usepackage{substitutefont}
  \substitutefont{T2A}{\familydefault}{cmr}
\fi

\usepackage[russian,british]{babel}
\usepackage{url}
\usepackage{pgf}

\usepackage{amsmath}

\usepackage{covington} 

\usepackage{hyperref}

\dialogfinalcopy 

\title{Is neural language acquisition similar to natural? \\ A chronological probing study}

\author{Ekaterina Voloshina \\
  AIRI, HSE University \\
  Moscow, Russia \\
  {\tt vokat@mail.ru} \\\And
  Oleg Serikov \\
  AIRI, DeepPavlov, HSE University \\
  Moscow, Russia \\
  {\tt srkvoa@gmail.com} \\\And
  Tatiana Shavrina \\
  AIRI, SberDevices \\
  Moscow, Russia \\
  {\tt rybolos@gmail.com} \\}

\date{}

\begin{document}
\maketitle
\begin{abstract}

The probing methodology allows one to obtain a partial representation of linguistic phenomena stored in the inner layers of the neural network, using external classifiers and statistical analysis.

Pre-trained transformer-based language models are widely used both for natural language understanding (NLU) and natural language generation (NLG) tasks making them most commonly used for downstream applications. However, little analysis was carried out, whether the models were pre-trained enough or contained knowledge correlated with linguistic theory.

We are presenting the chronological probing study of transformer English models such as MultiBERT and T5. We sequentially compare the information about the language learned by the models in the process of training on corpora. The results show that 1) linguistic information is acquired in the early stages of training 2) both language models demonstrate capabilities to capture various features from various levels of language, including morphology, syntax, and even discourse, while they also can inconsistently fail on tasks that are perceived as easy. 

We also introduce the open-source framework for chronological probing research, compatible with other transformer-based models. \url{https://github.com/EkaterinaVoloshina/chronological_probing}
  
  
  \textbf{DOI:} 10.28995/2075-7182-2022-20-XX-XX
\end{abstract}

\selectlanguage{russian}
\begin{center}
  \russiantitle{Усвоение языка у языковых моделей и человека: хронологическое пробинг-исследование. }

  \medskip \setlength\tabcolsep{0.5cm}
  \begin{tabular}{ccc}
    \textbf{Екатерина Волошина} & \textbf{Олег Сериков} & \textbf{Татьяна Шаврина}\\
      AIRI, НИУ ВШЭ & AIRI, DeepPavlov, НИУ ВШЭ & AIRI, SberDevices\\
      Москва, Россия & Москва, Россия & Москва, Россия\\
      {\tt vokat@mail.ru} &  {\tt srkvoa@gmail.com} & {\tt rybolos@gmail.com}
  \end{tabular}
  \medskip
\end{center}

\begin{abstract}
  
Пробинг-методология позволяет получить представление о явлениях языка, хранящееся во внутренних слоях нейросети, с помощью внешних классификаторов и статистического анализа.

Предобученные языковые модели на основе трансформерных архитектур широко используются как для задач понимания естественного языка (NLU), так и для задач генерации (NLG), что делает их наиболее часто используемыми для широкого ряда применений. Однако, недостаточно проводился анализ, достаточно ли предобучены модели и содержат ли знания, коррелирующие с теоретическими представлениями о языке.

Мы представляем исследование на основе хронологического пробинга на примере моделей MultiBERT и T5, в котором последовательно исследуем выучиваемую информацию о языке в процессе предобучения моделей на корпусе. Результаты показывают, что 1) лингвистическая информация усваивается уже на ранних этапах обучения 2) обе языковые модели демонстрируют способность фиксировать различные свойства языка на разных уровнях, включая морфологию, синтаксис и дискурс, в то же время они могут не справляться с задачами, которые воспринимается как простые.

Мы также предоставляем открытый фреймворк для хронологического пробинга, совместимый с языковыми моделями на основе архитектур transformer. \url{https://github.com/EkaterinaVoloshina/chronological_probing}
  
  \textbf{Ключевые слова:} пробинг, усвоение языка, языковые модели, трансформеры
\end{abstract}
\selectlanguage{british}
\section{Introduction}
\label{intro}

%
%

 
The role of deep learning language models has been increasing in the field of methodology for linguistic research, providing new methods for both diachronic and synchronic studies \cite{manning2015last}. 
In particular, transformer-based language modeling research has produced a variety of tools that may discover regularities and structures in data, many of which have resulted in practical applications. 

In this study, we search for a match between the language competencies of popular language models and compare their results with the levels of a first language learner. 
As the transformer models are expected to acquire a language during the training process, the probing methodology has shed light on model training success. Probing tasks are usually classification tasks where classes represent different values of a linguistic features, such as a subject number, tree depth, and a connector type. 
Theoretical representation of language often inquires about the levels of phonetics, morphology, syntax, and discourse/pragmatics to be involved in a probing study \footnote{However, some researchers \cite{embick2007distributed,caha2009nanosyntax} doubt that a language functions as a level system. They suggest that morphology and syntax operate at the same time. Other researchers argue that morphology and syntax are different layers of a language.}. 

The main focus of this work is to explore how language models acquire measurable linguistic structures during training. The contributions of our work are the following:
\begin{itemize}
\item We propose a methodology for chronological probing, based on checkpoint-wise result comparison during model training\footnote{ \url{https://github.com/EkaterinaVoloshina/chronological_probing}}. We denote chronological probing as any probing technique that refers to the training history/iterations of the same model.
\item We test two models (MultiBERT\cite{sellam2021multiberts} and T5\cite{raffel2019exploring}) on existing 12 different probing tasks in morphology, syntax, and discourse and present an analysis of the models' gradual learning of language phenomena, in comparison with the well-known facts about the acquisition of the first language by a child.  
\item We present the evaluation results for the named models and state that the models tend to learn the linguistic phenomena in a specific order, and some parts of grammar are ``acquired'' first.
\end{itemize}

The presented framework and methodology are available open-source under Apache 2.0 license.

\section{Related work}
Until recent years, the task of learning syntax, which every five-year-old child performs effortlessly, has eluded brute language modeling force. This makes the language models a particular object of study, considered both from the interpretability and modeling language acquisition. 
As \cite[p.~119]{de2011handbook} states, \textit{``computational models of language acquisition must begin and end as an integral part of the empirical study of child language.''} 

Following this thesis, we turn our attention to the probing methodology and comparable case studies in the field of language acquisition, focusing on the transformer architectures.
\subsection{Probing and approaches to the black box of language modeling}

An increasing number of works are devoted to interpreting language models from a linguistic point of view. 
The quickly advancing field of \textit{probing} received lots of researchers' attention when the hegemony of the large black-box models was set up.
Researchers question the extent of the models' ``understanding'' of the language in probing. They inspect if, and to what limits, the language models' behavior agrees with the insights of the theory of language.
Following the hierarchy of language levels (morphology, syntax, discourse)~\cite{dalrymple2001lexical}, the probing studies often suppose the experiments related to models' proficiency on a certain level of language.

This line of research typically comes down to analyzing how linguistic structures are represented in a model's knowledge. 
Such structures represent syntagmatic/paradigmatic mechanisms (how language units combine and alternate, respectively) of language.
It is believed \cite{mccoy-etal-2020-berts}, that rediscovering these structures would help models to get closer to human performance on a variety of tasks.
 
Probing, in general, considers how interpretable the behavior of the language model wrt the linguistic properties of the data.
A huge body of probing studies rely on linear models (e.g., external classifiers \cite{belinkov2016probing}) that try to establish the relationship between internal representations from the language model and the desired linguistic phenomena. Thus, the linear correlation is measured between the model's forward pass embeddings and the linguistic properties of the passed data.
A sample study \cite{Tenney2019} could measure the strength of correlation between a model's particular layer activations on some word and word's part-of-speech.


Strong correlations have been recorded when comparing the models' forward pass activations with the passed data underlying linguistic structure~\cite{Belinkov2018,Tenney2019,Conneau2018,hewitt-liang-2019-designing} using probing methods. 

Such a high performance could be misleading. The properties of the model and the properties of the used data impact the resulting score of the correlation probing study.
Thus, given only a correlation score, one does not know if it reflects the model's (but not the corpus itself) linguistic informativeness.
As a result, several approaches to conducting more reliable studies have been proposed. ~\cite{hewitt-liang-2019-designing,Zhang2018,voita2020information,Pimentel2020}. 

The probing methodology combining various annotated data is commonly used as the benchmark for language model comparison and evaluation of their generalizing ability. The SentEval toolkit \cite{conneau2018senteval} has led to the popularization of the 10 tasks used to distinguish between random and justified, brittle, and robust results of model training, including different types of architectures. However, analogous research on the same architecture or even the same model is in its early development stage.
The first work on probing of neural networks across time was carried by \cite{saphra2018understanding}. The authors showed that first, LSTM acquires syntactic and semantic features and later information structure. \cite{chiang2020pretrained} looked at the training process of ALBERT and concluded that semantic and syntactic information is acquired during the early steps while world knowledge fluctuates during the training. \cite{liu2021probing} showed similar results on RoBERTa: the model shows good results on linguistic probing tasks starting from early stages, and later it learns factual and commonsense knowledge. 

Chronological probing could enrich the interpretable documentation of model training in time and thus explore the new aspects of model training and more clearly expose its problems.

\subsection{Language acquisition and language models}
Language learning is one of the quintessential human traits. 
First language acquisition(LA), unites both neurocognitive research, psycholinguistics, and computational approaches, focusing on the ability to acquire the capacity to perceive and comprehend language.

\paragraph{Statistical language acquisition}
Language modeling has formed a branch in language acquisition studies named statistical language acquisition. Various aspects of language, including phonological, syntactic, lexical, morphological, and semantic features, were investigated in terms of statistical patterns children receive with the linguistic input. Recent studies postulating qualitative and quantitative measures of LA include: 
\begin{itemize}
\item \textbf{Morphology and Syntax} Morphology and syntax studies across language acquisition studies are definitely those explored the most. Starting with the poverty of stimulus problem and the argument between innateness and learning of grammar, it has led to typologically various sets of descriptive works and even computational models of the acquisition process. Thus, \cite{lewis2001learnability} train a simple RNN to discriminate between grammatical strings that follow the inversion rule and those that do not (e.g., moving the first auxiliary verb such as “Is the man that tall is nice?”). The training data for the study is generated artificially and fails to prove that such a network generalizes on a mixture of diverse syntactic constructions. \cite{reali2005uncovering} use bigram models to capture the patterns of auxiliary inversion based on lifelike data from child-directed speech. The model can consistently assign higher probabilities to grammatical strings than ungrammatical strings, which was interpreted as having successfully learned the correct inversion rule. However, as \cite{kam2008bigrams} note, this result is because bigrams such as “who are” are much more frequent than the ungrammatical strings.  \cite{prefors2006poverty} approaches the structure dependency problem with Bayesian learning and attempts to learn a grammar that could generate additional sentences. The model evaluates and selects between two grammars, a finite state grammar and a context-free grammar constructed by the authors based on a simplified subset of child-directed English.

It is worth noting how similar all the problem formulations are to the modern formulations of probing classification problems described below. They are also far from a complete description of the process of mastering grammar. 
    
\item \textbf{Discourse} The creation of texts, not sentences, with various discourse features, such as competence in speech acts, conversations, speech registers, and extended speaking turns, is more often considered a later stage of speech development. There are no computer models for the assimilation of discursive properties comparable to models for morphology and syntax. However, research in this direction is underway.  

\end{itemize}
 

In \cite{ororbia-oleg-viz}, authors examine whether neural language models acquire language better when trained in a multi-modal setting (namely, accompanied with visual contexts) compared to traditional purely textual pre-training. They show that indeed providing models with perceptual context is beneficial for training language models. Authors claim this evidence to correspond with the theory of situated cognition introduced in \cite{greeno1993situativity}.

In this work, we propose the first methodological step for chronological interpretation of traditional transformer language models in the framework of LA.

\section{Experimental setup}
\subsection{Models}
We calculated the accuracy of two transformer models on 12 probing tasks. As we want to know how universal patterns of language acquisition in models are, we experiment with two different transformer architectures: BERT and T5. While BERT has only encoder layers, T5 includes both encoder and decoder layers. Therefore, embeddings from BERT come from the encoder, and T5 embeddings are calculated after going through decoder after encoder. 

For this work, we use already published models with available checkpoints. It means that they were trained on different data and computational powers. Moreover, they were trained with different batch sizes (256 for MultiBERT and 32 for T5). However, we follow \cite{chiang-etal-2020-pretrained,liu2021probing} and measure the training progress in iterations. The further comparison of the two models is indicative only.

\textbf{MultiBERT} \cite{sellam2021multiberts} is based on BERT-base-uncased architecture, and it is the model of the same size (12 layers and embedding size 768). Unlike the original BERT\cite{devlin2019bert}, it was trained with 25 different seeds. The authors also released checkpoints of the first five models. We use the model with seed 0 in our experiments. MultiBERT was trained on both literary and non-fiction texts. The former comes from BookCorpus \cite{zhu2015aligning}, which includes 11,038 books of 16 different genres. The non-fiction texts are taken from English Wikipedia collected by \cite{turc2019well}.

\textbf{T5-small} model is trained within the \texttt{t5-experiments} framework.\footnote{\url{https://github.com/yurakuratov/t5-experiments}} and follows the Hugging Face implementation of T5 \cite{raffel2019exploring}. 
It consists of 6 layers with 512 embedding size.

Following the previous language, modeling works \cite{devlin2019bert,bojanowski2017enriching}, we use the Wikipedia data to train the model. The raw Wikipedia data is provided by The Pile project \cite{gao2020pile} contains $\approx 19 \text{Gb}$ of expository prose texts of various domains, and is treated as a language modeling dataset of reasonably well quality.

\textbf{Baseline}
As a baseline, we use the method described in \cite{hewitt2019designing}. We train logistic regression on top of embeddings of models mentioned above with shuffled class labels.

\begin{table}[]
\resizebox{\textwidth}{!}{
\begin{tabular}{@{}lp{10cm}l@{}}
\toprule
 Task &  Sentence examples & Labels  \\
 \midrule
 Subject number &	\textit{Her employer had escaped with his wife for several afternoons this summer.}  & NN \\ & 
 \textit{Your Mackenzie in-laws have sordid reputations few decent families wish to be connected with .} & NNS \\
 \midrule 
 Person &   \textit{So I still can recomend them but prepare pay twice as much as they tell you initially .} &  has a person marker \\ &  \textit{The service was friendly and fast , but this just does nt make up for the lack - luster product .} & does not have a person marker\\
 \midrule
Tree depth & \textit{We have done everything we can for her .} & 11 \\ & \textit{Alvin Yeung of Civic Party} & 3  \\
\midrule
Top constituents &  \textit{Did it belong to the owner of the house ?} &  VBD\_NP\_VP\_. \\ & \textit{How long before you leave us again ?} &  WHNP\_SQ\_. \\
\midrule
Connectors & \textit{He 'd almost forgotten about that man .	Sarah had somehow brought him back , just as she had his nightmares .} & but \\ & \textit{I let out a slow , careful breath .	Felt tears sting my eyes .} & and \\
\midrule
Sentence position &  \textit{Quneitra Governorate ( / ALA-LC : `` Muhafzat Al-Qunaytrah `` ) is one of the fourteen governorates ( provinces ) of Syria . The governorate had a population of 87,000 at the 2010 estimate . Its area varies , according to different sources , from 685 km ² to 1,861 km ² . It is situated in southern Syria , notable for the location of the Golan Heights . The governorate borders Lebanon , Jordan and Israel .} & 1 \\ & \textit{The bossom and the part of the xhubleta covered by the apron are made out of crocheted black wool . The bell shape is accentuated in the back part . The xhubleta is an undulating , bell-shaped folk skirt , worn by Albanian women . It usually is hung on the shoulders using two straps . Part of the Albanian traditional clothing it has 13 to 17 strips and 5 pieces of felt .} & 4 \\
\midrule
Penn Discourse Treebank & \textit{Solo woodwind players have to be creative,they want to work a lot} & Pragmatic Cause \\ & 
\textit{The U.S. , along with Britain and Singapore , left the agencyl, its anti-Western ideology , financial corruption and top leadership got out of hand} & List \\
\midrule
Discourse Coherence &  \textit{Within the fan inlet case , there are anti-icing air bosses and probes to sense the inlet pressure and temperature .', 'High speed center of pressure shifts along with fin aeroelasticity were major factors . At the 13th ( i.e .', 'the final ) compressor stage , air is bled out and used for anti-icing . The amount is controlled by the Pressure Ratio Bleed Control sense signal ( PRBC ) . The `` diffuser case `` at the aft end of the compressor houses the 13th stage .} & a text is not coherent \\ & 
\textit{This experience of digital circuitry and assembly language programming formed the basis of his book `` Code : The Hidden Language of Computer Hardware and Software '' . Petzold purchased a two-diskette IBM PC in 1984 for \$ 5,000 . This debt encouraged him to use the PC to earn some revenue so he wrote an article about ANSI.SYS and the PROMPT command . This was submitted to PC Magazine for which they paid \$ 800 . This was the beginning of Petzold 's career as a paid writer . In 1984 , PC Magazine decided to do a review of printers .} & a text is coherent \\
\bottomrule
\end{tabular}}
\caption{Examples of tasks}
\label{tab:tasks}
\end{table}

\begin{table}[]
\resizebox{\textwidth}{!}{
\begin{tabular}{@{}lp{10cm}p{10cm}@{}}
\toprule
Task & Acceptable sentence & Unacceptable sentence \\
\midrule
Transitive & \textit{The pedestrians question some people.} & 
\textit{The pedestrians wave some people.}  \\
\midrule
Passive & \textit{Tracy isn't fired by Jodi's daughter.} &  \textit{Tracy isn't muttered by Jodi's daughter.} \\
\midrule
Principle A c command & \textit{This lady who is healing Charles wasn't hiding herself.}  & 
\textit{This lady who is healing Charles wasn't hiding himself.}\\
\midrule
Adjunct Island  & \textit{Who does John leave while alarming Beverly?} & \textit{Who does John leave Beverly while alarming?} \\
\bottomrule
\end{tabular}}
\caption{BLiMP Minimal pairs examples}
\label{tab:blimp}
\end{table}

\subsection{Probing tasks}

Probing tasks come from several datasets published earlier: SentEval \cite{conneau2018probing}, Morph Call \cite{mikhailov-etal-2021-morph}, DisSent \cite{nie2019dissent}, DiscoEval \cite{mchen-discoeval-19}, and BLiMP \cite{warstadt2020blimp}. The class balance of first eight tasks is illustrated with Figure \ref{fig:class_balance} in Appendix. We choose these tasks to make our results comparable to other works on probing.

As we want to show another perspective on language acquisition, we balance classifier probing tasks with BLiMP tasks. As BLimP only covers morphology and syntax, all discourse-based tasks are evaluated with a classifier.

The datasets from Benchmark of minimal linguistic pairs (BLiMP) have a structure different from other tasks: every task includes pairs with minimal differences to illustrate one of the grammatical features of English. One sentence of the pair is grammatical, whereas another one is unacceptable. We chose four BLiMP tasks for our experiments: transitive and passive verbs, Principle A of C command, and Island effects. For the first two tasks, pairs have different verbs, where only one verb is transitive or can be used in a passive form. These tasks are categorized as morphological (see Table \ref{tab:blimp}). 

The second two tasks reflect syntactic effects on English. The Principle A task shows the use of reflexives. According to \cite{chomsky1981lectures}, a reflexive should have a local antecedent.

The task on Island effects tests a model's sensibility to syntactic order. Island is a structure from which a word can not be moved \cite{ross1967constraints}. The sentence is unacceptable if a word is moved out of an island.

The tasks from other datasets are summarized below:
\begin{itemize}
    \item \textbf{Subject number} (SentEval): this task is supposed to show how models acquire morphology. It is a binary classification task with labels NNS and NN (plural and singular number, respectively). The classifier should decide on a sentence class based on the number of sentence subjects. 
    \item \textbf{Person} (Morph Call): this task is also morphological. It is a binary classification with labels 0 and 1, which signifies if a subject has a person marker or not.
    \item \textbf{Tree depth} (SentEval): this task contains six classes, which stands for the depth of the syntactic tree of a given sentence. Hence, this task is meant to show the level of syntax acquisition. 
    \item \textbf{Top constituents} (SentEval): this multiclass task belongs to the group of syntactic tasks. The aim is to choose a class that includes constituents located right below the sentence (S) node.
    \item \textbf{Connectors} (DisSent): this dataset includes pairs of sentences originally connected with one of 5 prepositions, and the task is to choose the omitted preposition. It is supposed to show how models catch discourse relations.
    \item \textbf{Sentence position} (DiscoEval): this dataset contains sequences of 5 sentences, and the first sentence is placed in the wrong place. Therefore, the aim is to detect the original position of these sentences. This task is also meant to show models' accuracy in discourse.
    \item \textbf{Penn Discourse Treebank} (DiscoEval): the task is based on Penn Discourse Treebank annotation. The aim is to choose the right discourse relation between two discourse items from Penn Treebank. 
    \item \textbf{Discourse coherence} (DiscoEval): this task is a binary classification with classes 1 and 0. Class 1 means that the given paragraph is coherent, and class 0 should be assigned to paragraphs with shuffled sentences. 
\end{itemize}

\subsection{Probing Methods}
\textbf{Sentence embedding classification}:
Token embeddings from transformer models are turned into sentence embedding via mean pooling. Then logistic regression classifies embeddings' sentences. This method is used with tasks from SentEval and Morph Call.

\textbf{Positional sentence classification}:
For the Sentence Position task, we used the following method. First, we count sentence embeddings as described above. Then the difference between the first embedding and the other is calculated pair-wisely. The first embedding and its differences with others are concatenated and put as input to logistic regression.

\textbf{Sentence embedding concatenation \& classification}:
We concatenated sentence embeddings for other discourse tasks, which were calculated as the mean of token embeddings. The concatenated sentence embeddings served as inputs for logistic regression.

\textbf{Masking tasks}:
The probing task is based on the idea of masking language modeling. In a sentence, each word is masked, and then its probability is summed with other words' probabilities. The probability of an acceptable sentence should be higher than the probability of an unacceptable sentence. This method is for use for all tasks from BLiMP. 

\section{Results}

\subsection{Results of MultiBERT}

\begin{figure}[!h]
\includegraphics[width=\linewidth]{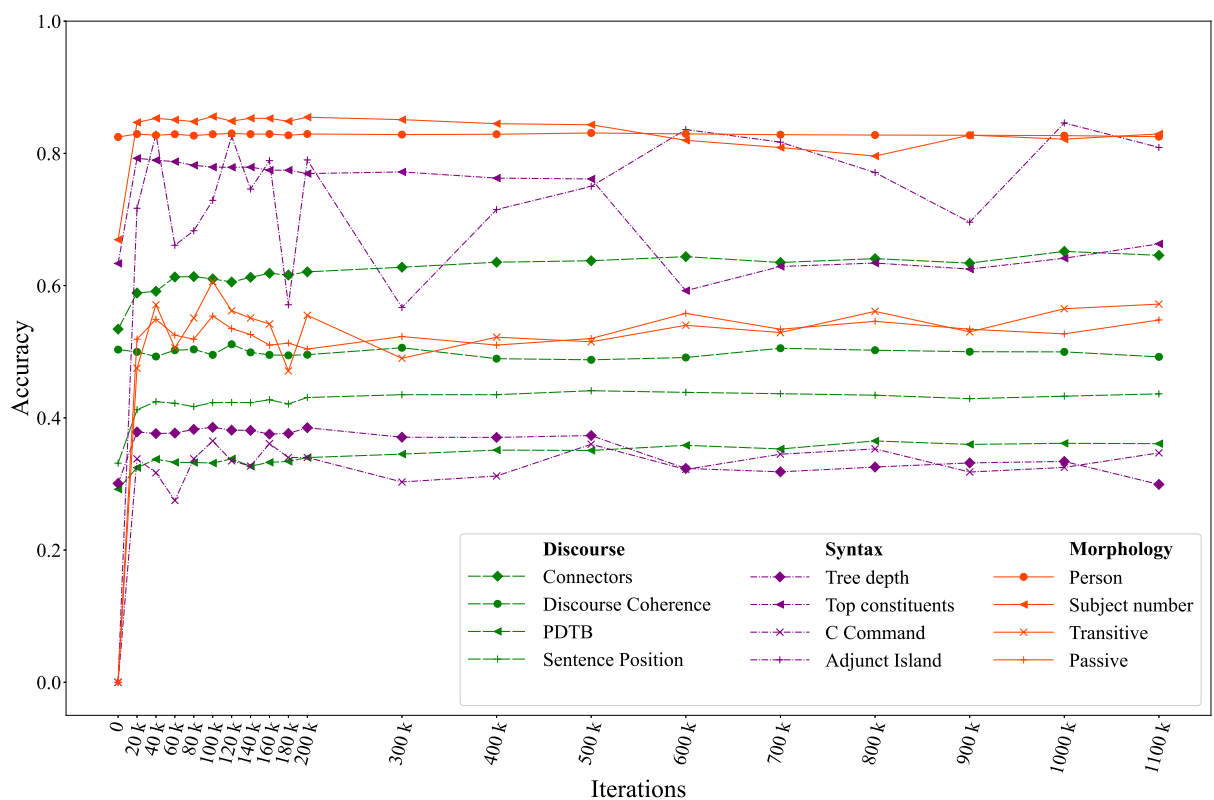}
\caption{Comparison of MultiBERT results on different tasks. How to read this figure: from left to right, on the X axis, we see results of intermediate evaluation on the task during model pre-training. Each iteration is equal to 25,600,000 sentences for MultiBERT and 3,200,000 sentences for T5. The Y-axis shows the accuracy metric on the tasks. Tasks are shown in the legend in different colors. As we can see, in the process of model pre-training, there already is a gradual increase in accuracy in tasks related to morphology (shown in orange) in the early stages. The information in the model embeddings stabilizes fairly quickly and remains stable from the 20,000th training step. The same can not be said for tasks related to syntax (shown in purple): their quality remains unstable and fluctuates quite a lot during pre-training. Discourse tasks (green) remain stable at a low-quality level from the start and tend to improve the metrics very slowly.
}
\centering
\label{fig:bert}
\end{figure}

The results of the experiments with the MultiBERT-base model are summarized in Figure \ref{fig:bert}. The model shows the best results on Subject Number and Person tasks. The classification of PDTB relations, Tree depth, and Principle A acceptability are performed with the worst accuracy.

As seen from Figure \ref{fig:bert}, accuracy of models stop changing after 600,000 iterations. However, there is a significant difference between tasks from BLiMP and other datasets. For example, the performance on the Adjunct Island task remains unstable during the whole period of iterations. 
Another difference between these tasks lies in the quality of the models. It is illustrated with tasks grouped as ``morphological'': Subject Number and Person tasks, which use logistic regression on MultiBERT embeddings, are solved much better than Transitive and Passive verbs. However, as follows from the plot, it is hard to group tasks based on the absolute value of accuracy. 

The changing dynamics provide another perspective. From this point of view, all tasks grouped as ``discourse'' show a similar feature: unlike others, their quality does not fluctuate but rather slightly grows across the training time. 
On other tasks, models increase the quality during the first iterations. ``Syntactic'' tasks tend to change even during later iterations. However, it is not a strict rule, and some tasks show similar behavior to ``morphological'' ones.

\subsection{Results of T5}
\begin{figure}[!h]
\includegraphics[width=\linewidth]{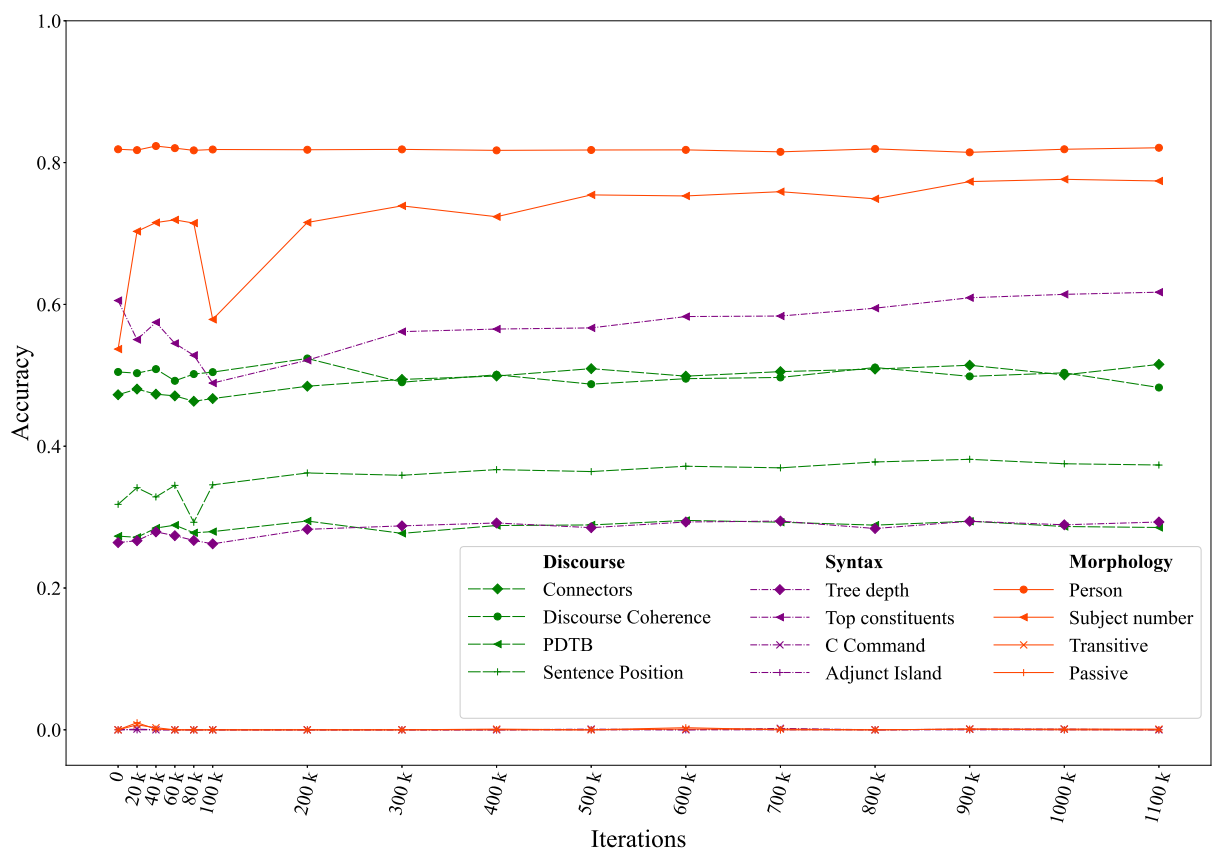}
\caption{Comparison of T5 results on a different task. How to read this figure: from left to right, on the X axis, we see results of intermediate evaluation on the task during model pre-training. Each iteration is equal to 25,600,000 sentences for MultiBERT and 3,200,000 sentences for T5. The Y-axis shows the accuracy metric on the tasks. Tasks are shown in the legend in different colors. As we can see, in the case of the T5 model, the task quality seems to be more stable from the beginning, with a few exceptions like subject number classification. Most of the tasks show the slow yet gradual growth of the metrics, but somehow not the verb transitivity classification.}
\centering
\label{fig:t5}
\end{figure}
Due to the architecture, the significant difference in T5 results is the zero-close quality on BLiMP datasets. Except for these tasks, the quality of T5 is similar to MultiBERT. The best performance is on the Person task, and the worst quality is shown on PDTB relation classification and Tree depth. 

Unlike MultiBERT, we first used the available checkpoints of T5 with a step of 100,000 iterations. Then we trained a new model on the same resources and texts, but it might have a better initialization, which affected the final results.

Similar to MultiBERT, discourse tasks show almost no significant change and slow growth, whereas the model increases its results on syntactic and morphological tasks during the first 100,000 iterations. 

\subsection{Comparison of models}
\begin{figure}[!h]
\includegraphics[width=16cm]{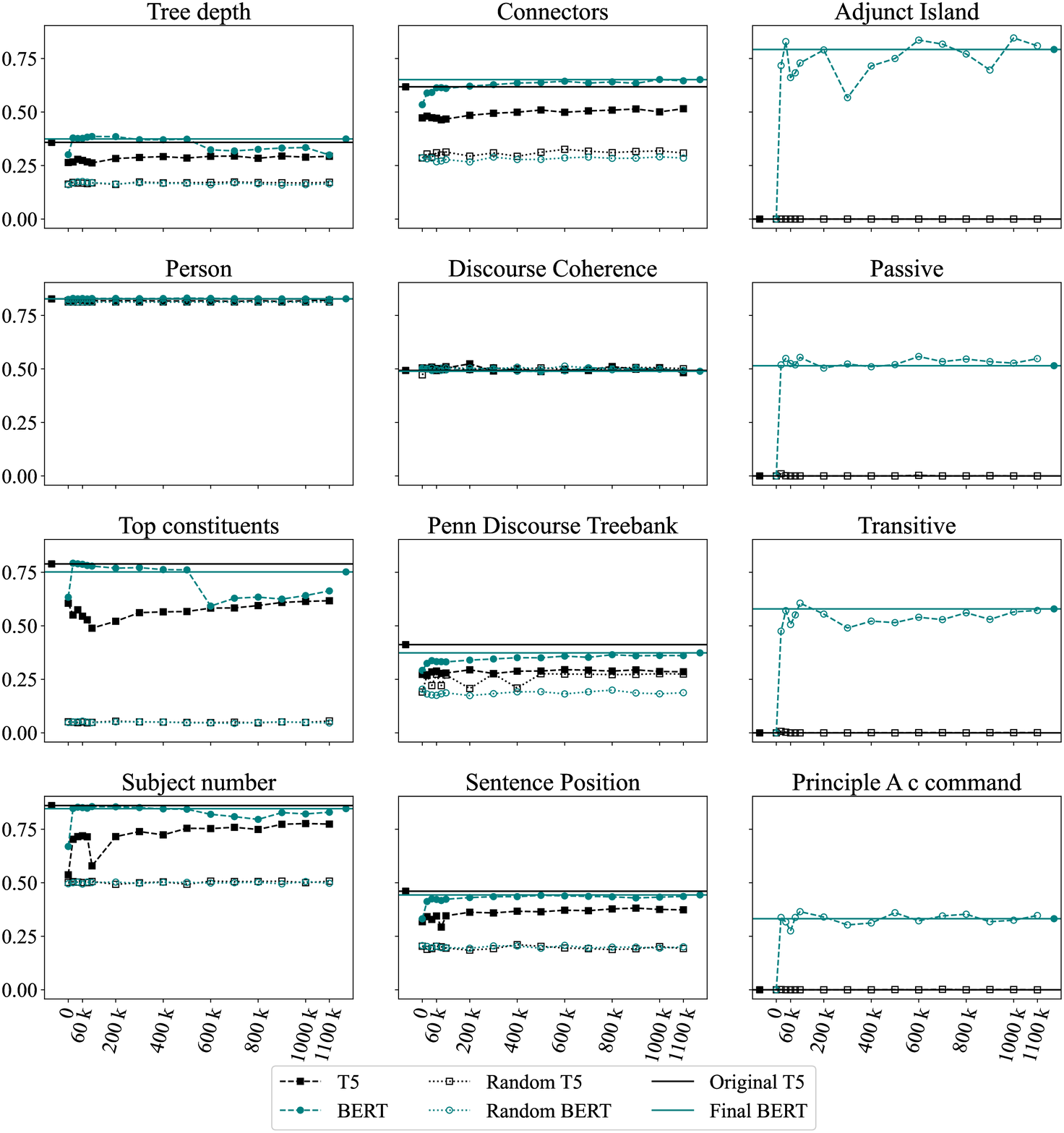}
\caption{Performance on models on different tasks. The detailed task-wise comparison shows the difference in T5 and MultiBERT training results. The models are compared to the final available checkpoints of the models (\textit{Original T5, Final BERT}) and with the random baseline.}
\centering
\label{fig:tasks}
\end{figure}

We described the surface results of models' performance and now can deep into more detailed results. The results described above should be considered relative. To illustrate how much information models acquire during these iterations, we compare them to final models. As the process of training T5 was not finished, we compared this model with the original T5. As seen from Figure \ref{fig:tasks}, MultiBERT scores are close to the results of the final checkpoint. Hence, there is no need to look at later iterations. The comparison with the original T5 shows that the model we use performs worse due to the smaller resources it was trained on. Therefore, the difference in quality should not be explained by the difference in architecture. 

However, we should consider that some tasks are performed with the same quality as embeddings with shuffled labels (Discourse coherence and Person). Moreover, T5 does not perform much better than the Penn Discourse Treebank relations baseline. Consequently, models encounter difficulty with discourse tasks.

Furthermore, MultiBERT and T5 show similar learning trajectories on several tasks, such as Connectors and Sentence Position tasks. 
Another key feature shared by the two models is the termination of increases between 500,000 and 600,000 iterations. 
Despite the fact that models vary in size and training process, they show some similarities in probing tasks. Hence, the acquisition generally does not depend on the model architecture.

\section{Discussion}

Our results show that linguistic information is acquired fast, before 600,000 training iterations. It corresponds to results of other researchers  \cite{chiang2020pretrained,liu2021probing} that independently showed similar results on a fast acquisition of linguistic features. However, discourse is not fully acquired by the end of the observed training period compared to the baseline results. The difference between syntactic and morphological tasks is insignificant. It correlates with ideas in morphosyntax. Although we can not prove that morphology and syntax are regarded as the same layer in models, we can make a less strict statement that models acquire all grammatical units simultaneously.


BLiMP gives another perspective on the process of acquisition. MultiBERT results remain unstable for a longer period than similar tasks with classifiers. It might indicate the difference between two different approaches to probing. However, from the linguistic point of view, BLiMP includes more difficult linguistic feature cases, while SentEval tasks test more basic knowledge. Hence, it could explain worse results. 

T5 architecture does not allow to use of this dataset in the same way as for MultiBERT since Masking Language Modeling and T5 generation are different tasks. We leave for further research an adaptation of this dataset for T5. 



\subsection{Human language acquisition results}
Many of the linguistic features used in probing tasks have been well studied in terms of their promptness and ease of acquisition by English speakers. First of all, it concerns morphology and syntax. Markers such as a person, number, and gender are acquired very early by children: before age two \cite{clark2017morphology}. Of course, in languages besides English, the acquisition of these features varies: if the feature is marked consistently with one affix and no morphonological alternation, children seem to find it easier to acquire. It is shown that the earlier mastery of case marking is present in languages like Hungarian and Turkish but not in German or Serbo-Croatian \cite{slobin1985crosslinguistic}.

Discursive features are acquired by children much later. Studies like \cite{pearson2003language} show that child's texts become more complex and decontextualized with age. Also, texts produced by children gradually progress in achieving more cohesion through ``referential and semantic links that bridge across sentences; they achieve coherence through a global hierarchical structure''. The discourse in these conversations between toddlers is tentative when neither side can be reliably significant. A longitudinal study of dialogues between two little girls aged 4 to 6, \cite{mctear1985children} traced the emergence of more and greater thematic continuity in their conversation as utterances began to play the dual role of responding to the preceding utterance as well as enabling further conversation. However, \cite{dorval1984developmental} showed that second-graders (eight-year-olds) were almost as likely to give unconditioned responses as conditionals, with no significant improvement seen until fifth grade.

Regarding the requirements that the sphere of language acquisition imposes on children, one can very carefully assess the limit in which the language models under consideration lie in terms of their abilities: their embeddings correspond to the level of language proficiency in a child under 11 years old.
\section{Conclusion}
Encoder and encoder-decoder language models have increased importance in tasks requiring understanding the natural language. The probing methodology we presented allows analyzing the changes within the language model during training, from epoch to epoch. The overall results of the work show that:

\begin{itemize}
 \item 
 T5 does not give any results on BLiMP due to the generation algorithm. Most tasks show that T5 acquires basic morphological and syntactic features and some discourse features.
  \item 
  MultiBERT shows results close to the trained model starting from 100,000 iterations. MultiBERT does not improve its quality on some discourse tasks compared to randomly labeled embeddings. However, it could be said that MultiBERT acquires each level to some extent.
 \item Both T5 and MultiBERT demonstrate comparable results regarding the quality of the language level acquisition. As we can not distinguish the factors between these results (whether this is the model's architecture, the training corpora, or both), we present these results 'as is' for the researchers that use them in the downstream tasks.
 \item Recording such results during training could save a lot of computational resources and time for interpreting the results, including downstream tasks. There are understandable context length limitations that prevent, for example, learning the discourse tasks. However, the results of the T5 model compared to the random embeddings on some tasks were unexpected or lower than expected.
  \item As the results show, the easiest tasks for the models tend to be morphology and syntax-related. These language level results are correlated and show a similar learning trajectory. Unlike morphology and syntax tasks, results on discourse-based tasks tend to be low, therefore, there is not enough evidence to claim that discourse has been learned.
  \item Using language acquisition as a trace can benefit in comparing general human language ability and modern language modeling methods. Drawing a border from above on the results on discursive tasks, we can say that in the current realities, the models do not pass the bar that 8-year-old children pass. 
\end{itemize}

We welcome both NLP and language acquisition research communities to reproduce the experimental setup and use the presented approach while training other architectures and contribute to formulating the better and more complex tasks correlated with language learning.

\bibliography{dialogue,custom,anthology}

\begin{thebibliography}{}

\bibitem[\protect\citename{Belinkov \bgroup et al.\egroup }2017]{Belinkov2018}
Yonatan Belinkov, Llu{\'\i}s M{\`a}rquez, Hassan Sajjad, Nadir Durrani, Fahim
  Dalvi, and James Glass.
\newblock 2017.
\newblock Evaluating layers of representation in neural machine translation on
  part-of-speech and semantic tagging tasks.
\newblock  // {\em Proceedings of the Eighth International Joint Conference on
  Natural Language Processing (Volume 1: Long Papers)}, P 1--10, Taipei,
  Taiwan, November. Asian Federation of Natural Language Processing.

\bibitem[\protect\citename{Belinkov}2016]{belinkov2016probing}
Yonatan Belinkov.
\newblock 2016.
\newblock Probing classifiers: Promises, shortcomings, and advances.
\newblock {\em Computational Linguistics}, P 1--12.

\bibitem[\protect\citename{Bojanowski \bgroup et al.\egroup
  }2017]{bojanowski2017enriching}
Piotr Bojanowski, Edouard Grave, Armand Joulin, and Tomas Mikolov.
\newblock 2017.
\newblock Enriching word vectors with subword information.
\newblock {\em Transactions of the Association for Computational Linguistics},
  5:135--146.

\bibitem[\protect\citename{Caha}2009]{caha2009nanosyntax}
Pavel Caha.
\newblock 2009.
\newblock The nanosyntax of case.

\bibitem[\protect\citename{Chen \bgroup et al.\egroup
  }2019]{mchen-discoeval-19}
Mingda Chen, Zewei Chu, and Kevin Gimpel.
\newblock 2019.
\newblock Evaluation benchmarks and learning criteria for discourse-aware
  sentence representations.
\newblock  // {\em Proc. of {EMNLP}}.

\bibitem[\protect\citename{Chiang \bgroup et al.\egroup
  }2020a]{chiang2020pretrained}
Cheng-Han Chiang, Sung-Feng Huang, and Hung-yi Lee.
\newblock 2020a.
\newblock Pretrained language model embryology: The birth of albert.
\newblock {\em arXiv preprint arXiv:2010.02480}.

\bibitem[\protect\citename{Chiang \bgroup et al.\egroup
  }2020b]{chiang-etal-2020-pretrained}
Cheng-Han Chiang, Sung-Feng Huang, and Hung-yi Lee.
\newblock 2020b.
\newblock {P}retrained language model embryology: {T}he birth of {ALBERT}.
\newblock  // {\em Proceedings of the 2020 Conference on Empirical Methods in
  Natural Language Processing (EMNLP)}, P 6813--6828, Online, November.
  Association for Computational Linguistics.

\bibitem[\protect\citename{Chomsky}1981]{chomsky1981lectures}
Noam Chomsky.
\newblock 1981.
\newblock Lectures on government and binding (dordrecht: Foris).
\newblock {\em Studies in generative grammar}, 9.

\bibitem[\protect\citename{Clark}2017]{clark2017morphology}
Eve~V Clark.
\newblock 2017.
\newblock Morphology in language acquisition.
\newblock {\em The handbook of morphology}, P 374--389.

\bibitem[\protect\citename{Conneau and Kiela}2018]{conneau2018senteval}
Alexis Conneau and Douwe Kiela.
\newblock 2018.
\newblock Senteval: An evaluation toolkit for universal sentence
  representations.
\newblock {\em arXiv preprint arXiv:1803.05449}.

\bibitem[\protect\citename{Conneau \bgroup et al.\egroup }2018a]{Conneau2018}
Alexis Conneau, German Kruszewski, Guillaume Lample, Lo{\"\i}c Barrault, and
  Marco Baroni.
\newblock 2018a.
\newblock What you can cram into a single {\$}{\&}!{\#}* vector: Probing
  sentence embeddings for linguistic properties.
\newblock  // {\em Proceedings of the 56th Annual Meeting of the Association
  for Computational Linguistics (Volume 1: Long Papers)}, P 2126--2136,
  Melbourne, Australia, July. Association for Computational Linguistics.

\bibitem[\protect\citename{Conneau \bgroup et al.\egroup
  }2018b]{conneau2018probing}
Alexis Conneau, German Kruszewski, Guillaume Lample, Loic Barrault, and Marco
  Baroni.
\newblock 2018b.
\newblock What you can cram into a single vector: Probing sentence embeddings
  for linguistic properties.
\newblock {\em arXiv preprint arXiv:1805.01070}.

\bibitem[\protect\citename{Dalrymple}2001]{dalrymple2001lexical}
Mary Dalrymple.
\newblock 2001.
\newblock {\em Lexical functional grammar}.
\newblock Brill.

\bibitem[\protect\citename{De~Villiers and Roeper}2011]{de2011handbook}
Jill~G De~Villiers and Thomas Roeper.
\newblock 2011.
\newblock {\em Handbook of generative approaches to language acquisition},
  volume~41.
\newblock Springer.

\bibitem[\protect\citename{Devlin \bgroup et al.\egroup }2019]{devlin2019bert}
Jacob Devlin, Ming-Wei Chang, Kenton Lee, and Kristina Toutanova.
\newblock 2019.
\newblock Bert: Pre-training of deep bidirectional transformers for language
  understanding.

\bibitem[\protect\citename{Dorval \bgroup et al.\egroup
  }1984]{dorval1984developmental}
Bruce Dorval, Carol~O Eckerman, and Susan Ervin-Tripp.
\newblock 1984.
\newblock Developmental trends in the quality of conversation achieved by small
  groups of acquainted peers.
\newblock {\em Monographs of the Society for Research in Child Development}, P
  1--91.

\bibitem[\protect\citename{Embick and Noyer}2007]{embick2007distributed}
David Embick and Rolf Noyer.
\newblock 2007.
\newblock Distributed morphology and the syntax/morphology interface.
\newblock {\em The Oxford handbook of linguistic interfaces}, 289324.

\bibitem[\protect\citename{Gao \bgroup et al.\egroup }2020]{gao2020pile}
Leo Gao, Stella Biderman, Sid Black, Laurence Golding, Travis Hoppe, Charles
  Foster, Jason Phang, Horace He, Anish Thite, Noa Nabeshima, et~al.
\newblock 2020.
\newblock The pile: An 800gb dataset of diverse text for language modeling.
\newblock {\em arXiv preprint arXiv:2101.00027}.

\bibitem[\protect\citename{Greeno and Moore}1993]{greeno1993situativity}
James~G Greeno and Joyce~L Moore.
\newblock 1993.
\newblock Situativity and symbols: Response to vera and simon.

\bibitem[\protect\citename{Hewitt and Liang}2019a]{hewitt-liang-2019-designing}
John Hewitt and Percy Liang.
\newblock 2019a.
\newblock Designing and interpreting probes with control tasks.
\newblock  // {\em Proceedings of the 2019 Conference on Empirical Methods in
  Natural Language Processing and the 9th International Joint Conference on
  Natural Language Processing (EMNLP-IJCNLP)}, P 2733--2743, Hong Kong, China,
  November. Association for Computational Linguistics.

\bibitem[\protect\citename{Hewitt and Liang}2019b]{hewitt2019designing}
John Hewitt and Percy Liang.
\newblock 2019b.
\newblock Designing and interpreting probes with control tasks.
\newblock {\em arXiv preprint arXiv:1909.03368}.

\bibitem[\protect\citename{Kam \bgroup et al.\egroup }2008]{kam2008bigrams}
Xu{\^a}n-Nga~Cao Kam, Iglika Stoyneshka, Lidiya Tornyova, Janet~D Fodor, and
  William~G Sakas.
\newblock 2008.
\newblock Bigrams and the richness of the stimulus.
\newblock {\em Cognitive science}, 32(4):771--787.

\bibitem[\protect\citename{Lewis and Elman}2001]{lewis2001learnability}
John~D Lewis and Jeffrey~L Elman.
\newblock 2001.
\newblock Learnability and the statistical structure of language: Poverty of
  stimulus arguments revisited.
\newblock  // {\em Proceedings of the 26th annual Boston University conference
  on language development}, volume~1, P 359--370. Citeseer.

\bibitem[\protect\citename{Liu \bgroup et al.\egroup }2021]{liu2021probing}
Leo~Z Liu, Yizhong Wang, Jungo Kasai, Hannaneh Hajishirzi, and Noah~A Smith.
\newblock 2021.
\newblock Probing across time: What does roberta know and when?
\newblock {\em arXiv preprint arXiv:2104.07885}.

\bibitem[\protect\citename{Manning}2015]{manning2015last}
Christopher~D Manning.
\newblock 2015.
\newblock Last words: Computational linguistics and deep learning.

\bibitem[\protect\citename{McCoy \bgroup et al.\egroup
  }2020]{mccoy-etal-2020-berts}
R.~Thomas McCoy, Junghyun Min, and Tal Linzen.
\newblock 2020.
\newblock {BERT}s of a feather do not generalize together: Large variability in
  generalization across models with similar test set performance.
\newblock  // {\em Proceedings of the Third BlackboxNLP Workshop on Analyzing
  and Interpreting Neural Networks for NLP}, P 217--227, Online, November.
  Association for Computational Linguistics.

\bibitem[\protect\citename{McTear}1985]{mctear1985children}
Michael McTear.
\newblock 1985.
\newblock {\em Children's conversation}.
\newblock B. Blackwell.

\bibitem[\protect\citename{Mikhailov \bgroup et al.\egroup
  }2021]{mikhailov-etal-2021-morph}
Vladislav Mikhailov, Oleg Serikov, and Ekaterina Artemova.
\newblock 2021.
\newblock Morph call: Probing morphosyntactic content of multilingual
  transformers.
\newblock  // {\em Proceedings of the Third Workshop on Computational Typology
  and Multilingual NLP}, P 97--121, Online, June. Association for Computational
  Linguistics.

\bibitem[\protect\citename{Nie \bgroup et al.\egroup }2019]{nie2019dissent}
Allen Nie, Erin Bennett, and Noah Goodman.
\newblock 2019.
\newblock Dissent: Learning sentence representations from explicit discourse
  relations.
\newblock  // {\em Proceedings of the 57th Annual Meeting of the Association
  for Computational Linguistics}, P 4497--4510.

\bibitem[\protect\citename{Ororbia \bgroup et al.\egroup
  }2019]{ororbia-oleg-viz}
Alexander Ororbia, Ankur Mali, Matthew Kelly, and David Reitter.
\newblock 2019.
\newblock Like a baby: Visually situated neural language acquisition.
\newblock  // {\em Proceedings of the 57th Annual Meeting of the Association
  for Computational Linguistics}, P 5127--5136, Florence, Italy, July.
  Association for Computational Linguistics.

\bibitem[\protect\citename{Pearson}2003]{pearson2003language}
Barbara~Zurer Pearson.
\newblock 2003.
\newblock Language acquisition: Discourse, narrative and pragmatics.
\newblock {\em Disertasi. USA: Department of Communication Disorders}.

\bibitem[\protect\citename{Pimentel \bgroup et al.\egroup }2020]{Pimentel2020}
Tiago Pimentel, Josef Valvoda, Rowan~Hall Maudslay, Ran Zmigrod, Adina
  Williams, and Ryan Cotterell.
\newblock 2020.
\newblock Information-theoretic probing for linguistic structure.
\newblock {\em CoRR}, abs/2004.03061.

\bibitem[\protect\citename{Prefors \bgroup et al.\egroup
  }2006]{prefors2006poverty}
Amy Prefors, Terry Regier, and Joshua~B Tenenbaum.
\newblock 2006.
\newblock Poverty of the stimulus? a rational approach.
\newblock  // {\em Proceedings of the Annual Meeting of the Cognitive Science
  Society}, volume~28.

\bibitem[\protect\citename{Raffel \bgroup et al.\egroup
  }2019]{raffel2019exploring}
Colin Raffel, Noam Shazeer, Adam Roberts, Katherine Lee, Sharan Narang, Michael
  Matena, Yanqi Zhou, Wei Li, and Peter~J Liu.
\newblock 2019.
\newblock Exploring the limits of transfer learning with a unified text-to-text
  transformer.
\newblock {\em arXiv preprint arXiv:1910.10683}.

\bibitem[\protect\citename{Reali and Christiansen}2005]{reali2005uncovering}
Florencia Reali and Morten~H Christiansen.
\newblock 2005.
\newblock Uncovering the richness of the stimulus: Structure dependence and
  indirect statistical evidence.
\newblock {\em Cognitive Science}, 29(6):1007--1028.

\bibitem[\protect\citename{Ross}1967]{ross1967constraints}
John~Robert Ross.
\newblock 1967.
\newblock Constraints on variables in syntax.

\bibitem[\protect\citename{Saphra and Lopez}2018]{saphra2018understanding}
Naomi Saphra and Adam Lopez.
\newblock 2018.
\newblock Understanding learning dynamics of language models with svcca.
\newblock {\em arXiv preprint arXiv:1811.00225}.

\bibitem[\protect\citename{Sellam \bgroup et al.\egroup
  }2021]{sellam2021multiberts}
Thibault Sellam, Steve Yadlowsky, Jason Wei, Naomi Saphra, Alexander D'Amour,
  Tal Linzen, Jasmijn Bastings, Iulia Turc, Jacob Eisenstein, Dipanjan Das,
  et~al.
\newblock 2021.
\newblock The multiberts: Bert reproductions for robustness analysis.
\newblock {\em arXiv preprint arXiv:2106.16163}.

\bibitem[\protect\citename{Slobin}1985]{slobin1985crosslinguistic}
Dan~I Slobin.
\newblock 1985.
\newblock Crosslinguistic evidence for the language-making capacity.
\newblock {\em The crosslinguistic study of language acquisition}, 2:1157--249.

\bibitem[\protect\citename{{Tenney} \bgroup et al.\egroup }2019]{Tenney2019}
Ian {Tenney}, Patrick {Xia}, Berlin {Chen}, Alex {Wang}, Adam {Poliak},
  R~Thomas {McCoy}, Najoung {Kim}, Benjamin {Van Durme}, Samuel~R. {Bowman},
  Dipanjan {Das}, and Ellie {Pavlick}.
\newblock 2019.
\newblock {What do you learn from context? Probing for sentence structure in
  contextualized word representations}.
\newblock {\em arXiv e-prints}, P arXiv:1905.06316, May.

\bibitem[\protect\citename{Turc \bgroup et al.\egroup }2019]{turc2019well}
Iulia Turc, Ming-Wei Chang, Kenton Lee, and Kristina Toutanova.
\newblock 2019.
\newblock Well-read students learn better: On the importance of pre-training
  compact models.
\newblock {\em arXiv preprint arXiv:1908.08962}.

\bibitem[\protect\citename{Voita and Titov}2020]{voita2020information}
Elena Voita and Ivan Titov.
\newblock 2020.
\newblock Information-theoretic probing with minimum description length.
\newblock {\em arXiv preprint arXiv:2003.12298}.

\bibitem[\protect\citename{Warstadt \bgroup et al.\egroup
  }2020]{warstadt2020blimp}
Alex Warstadt, Alicia Parrish, Haokun Liu, Anhad Mohananey, Wei Peng, Sheng-Fu
  Wang, and Samuel~R Bowman.
\newblock 2020.
\newblock Blimp: The benchmark of linguistic minimal pairs for english.
\newblock {\em Transactions of the Association for Computational Linguistics},
  8:377--392.

\bibitem[\protect\citename{Zhang and Bowman}2018]{Zhang2018}
Kelly Zhang and Samuel Bowman.
\newblock 2018.
\newblock Language modeling teaches you more than translation does: Lessons
  learned through auxiliary syntactic task analysis.
\newblock  // {\em Proceedings of the 2018 {EMNLP} Workshop {B}lackbox{NLP}:
  Analyzing and Interpreting Neural Networks for {NLP}}, P 359--361, Brussels,
  Belgium, November. Association for Computational Linguistics.

\bibitem[\protect\citename{Zhu \bgroup et al.\egroup }2015]{zhu2015aligning}
Yukun Zhu, Ryan Kiros, Rich Zemel, Ruslan Salakhutdinov, Raquel Urtasun,
  Antonio Torralba, and Sanja Fidler.
\newblock 2015.
\newblock Aligning books and movies: Towards story-like visual explanations by
  watching movies and reading books.
\newblock  // {\em Proceedings of the IEEE international conference on computer
  vision}, P 19--27.

\end{thebibliography}
\bibliographystyle{dialogue}

\section*{Appendix}
\begin{figure}[!h]
\includegraphics[width=\linewidth]{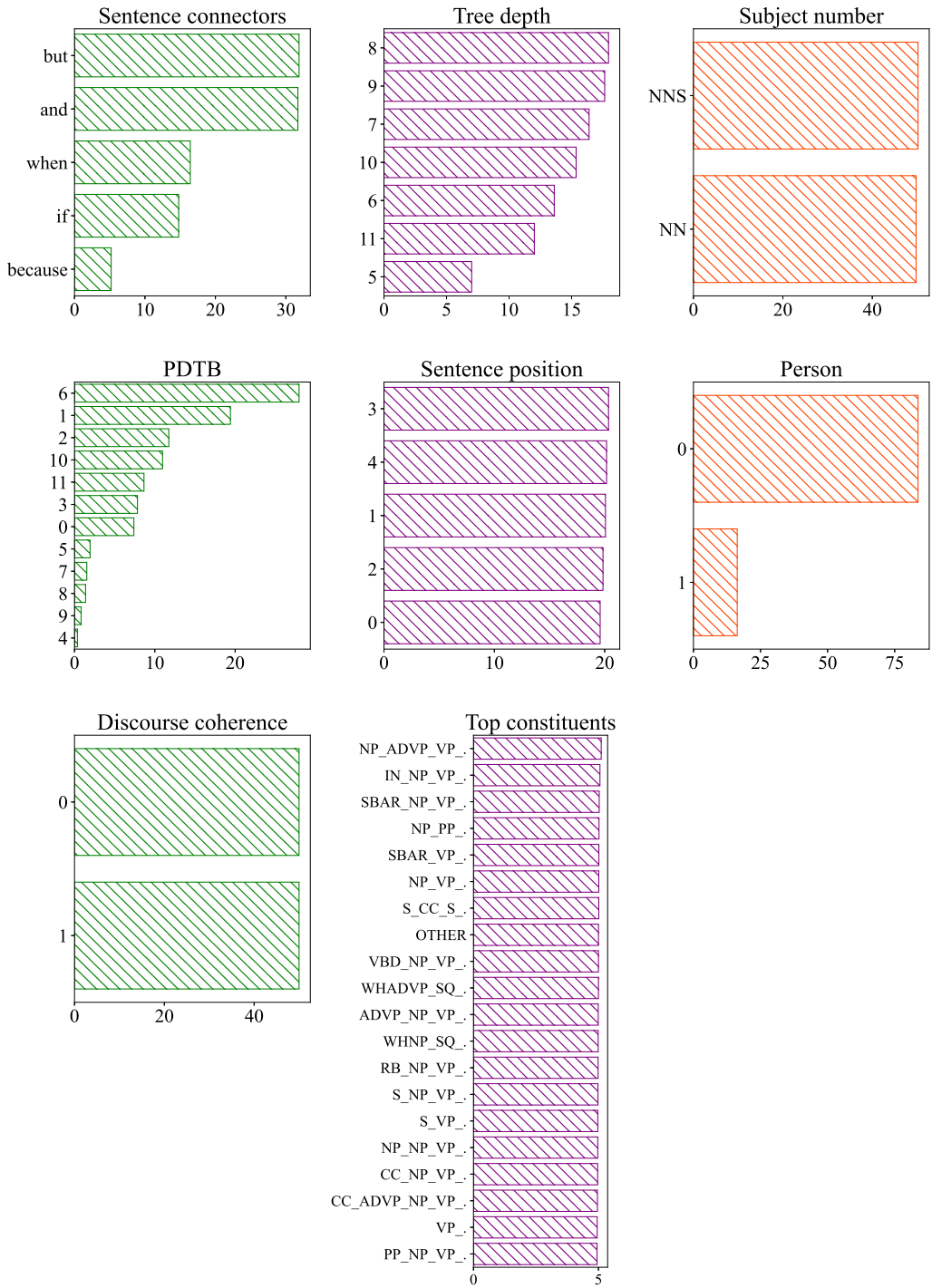}
\caption{The class balance of datasets}
\centering
\label{fig:class_balance}
\end{figure}



\end{document}